\documentclass{article}
\usepackage{stywhispers,amsmath,epsfig}

\usepackage{hyperref}
\usepackage{url}
\usepackage{times}
\usepackage{amsmath}
\usepackage{mathtools}
\usepackage{mathpazo}
\usepackage{multirow}
\usepackage{graphicx}
\usepackage{subfig}
\usepackage{booktabs}
\usepackage{epstopdf}
\usepackage{siunitx}

\title{Multitask learning of vegetation biochemistry from hyperspectral data}
\name{Utsav B. Gewali~\textsuperscript{1} and Sildomar T. Monteiro~\textsuperscript{1,2}}
\address{\textsuperscript{1}Chester F. Carlson Center for Imaging Science\\
		 \textsuperscript{2}Department of Electrical Engineering\\
		 Rochester Institute of Technology, Rochester, NY\\
		 ubg9540@rit.edu}
\begin{document}
%\ninept
%
\maketitle
\begin{abstract}
Statistical models have been successful in accurately estimating the biochemical contents of vegetation from the reflectance spectra.  However, their performance deteriorates when there is a scarcity of sizable amount of ground truth data for modeling the complex non-linear relationship occurring between the spectrum and the biochemical quantity. We propose a novel Gaussian process based multitask learning method for improving the prediction of a biochemical through the transfer of knowledge from the learned models for predicting related biochemicals. This method is most advantageous when there are few ground truth data for the biochemical of interest, but plenty of ground truth data for related biochemicals. The proposed multitask Gaussian process hypothesizes that the inter-relationship between the biochemical quantities is better modeled by using a combination of two or more covariance functions and inter-task correlation matrices. In the experiments, our method outperformed the current methods on two real-world datasets.

\end{abstract}
\section{Introduction}
\label{sec:intro}
Biochemistry prediction is the problem of estimating the contents of chemicals in vegetation from the measured reflectance spectrum. The presence of a certain chemical is manifested in the reflectance spectra as spectral absorption features, and the depth of the spectral feature is correlated to the contents of that chemical. It is, hence, possible to develop regression models to predict the contents of biochemicals from the reflectance spectrum~\cite{kokaly1999}. Recently, Gaussian processes have been successfully applied for this purpose~\cite{valls2009,verrelst2012,verrelst2013}. Gaussian processes, being non-parametric, can model the complex non-linear relationship that exists between the spectrum and the biochemical quantity very well. But, similar to other statistical models used for biochemistry prediction~\cite{bazi2007b}, they suffer in performance due to insufficient availability of training examples. 

Biochemistry prediction datasets commonly have very few ground truth data, due to the difficulties in the collection and the cost of chemical analysis of the vegetation samples. Moreover, many times, a single dataset may have few ground truth for a biochemical, while having plenty ground truth for other biochemicals. In this case, the task of predicting the contents of a biochemical may significantly benefit from the information learned from the task of predicting the contents of other related biochemicals. Multitask learning is the idea of learning a set of related tasks simultaneously such that the inter-relationship between the tasks can be exploited to improve the performance of each task. It is called transfer learning when the objective is to improve the performance of only a subset of tasks. Transfer and multitask learning have seen successful applications in many domains such as natural language processing~\cite{pan2010}, computer vision~\cite{lapin2014} and biomedical engineering~\cite{xu2011}. In this paper, we propose a novel multitask Gaussian process, inspired by \cite{bonilla2008}, with the motivation of improving the predictive performance of a biochemical, for which few training examples are available, using the information from related biochemicals. 

We evaluate our method with experiments on two real world datasets, and compare the results with the results from two state-of-the-art Gaussian process based multitask methods~\cite{bonilla2008,rakitsch2013}. In \cite{bonilla2008}, a task correlation matrix, learned from the data itself, is used to define a shared covariance representation over the tasks.  Our method extends \cite{bonilla2008} by using a combination of several task correlation matrices and covariance functions, instead of using a single task correlation matrix and a single covariance function, to model the covariance between the tasks. We postulate that our formulation gives the model more flexibility to learn the relationships between the tasks, and hence improves the prediction. The second method~\cite{rakitsch2013} is similar to \cite{bonilla2008}, except for that it also models a noise correlation matrix between the tasks. Modeling noise correlation helps to account for hidden sources effecting the tasks that are not included in the input. Our method could also be extended to include correlated noise.  This paper is organized as follows. Section \ref{sec:GP} provides a brief background on Gaussian process regression, Section \ref{sec:format} introduces the proposed method, Section \ref{sec:experiments} provides the evaluation of the proposed method on real datasets, and Section \ref{sec:discussion} discusses the implications of this study. 

\section{Gaussian Processes for Regression}
\label{sec:GP}
Gaussian process (GP) regression is a probabilistic model where the output variable values at all the training and the testing data points are considered to be samples of a joint multivariate normal distribution, having the mean vector zero and the covariance matrix given by a covariance function
~\cite{rasmussen2005}.  Inference about the posterior distribution of the output values at the test data points, which is also a multivariate normal distribution, is made by conditioning the joint normal distribution by the output values at the training points, and is given by  
\begin{equation}
\label{eq:gppred}
	f_*|X,y,X_* \, \sim \, \mathcal{N}(\bar{f_*}, cov(f_*)),
\end{equation}
where
\begin{align}
   \label{eq:gppredpara1}
  \bar{f_*} =& K(X_*,X) \, {[ K(X,X)+\sigma_n^2 I]}^{-1}y,  \\
  cov(f_*) =& K(X_*,X_*) - \nonumber \\
  \label{eq_gppredpara2}
  &K(X_*,X){[K(X,X)+\sigma_n^2 I]}^{-1} K(X,X_*),
\end{align}
  
and $f_*$ is a vector of output values at the test data points stored in the rows of the matrix $X_*$, and  $y$ is a vector of observed output values at the training data points stored in the rows of the matrix $X$.  $\sigma_n$ is the independent and identically distributed Gaussian noise variance observed at the output, and $K(X,X')$ is a covariance matrix whose \textit{i}-th row and \textit{j}-th column element is the $k(\mathbf{x},\mathbf{x'})$ of \textit{i}-th row of $X$ and \textit{j}-th row of $X'$. $k(\mathbf{x},\mathbf{x'})$ is a covariance function. It is usually parameterized by few free hyperparameters, which are learned from the data, along with  $\sigma_n$, by maximizing the log marginal likelihood function of the GP.

\section{Multitask learning with composite covariance function}
\label{sec:format}

We propose a new method to extend the method by Bonilla et al.~\cite{bonilla2008} and call it multitask learning with composite covariance function (MTGP-COMP). In \cite{bonilla2008}, the relationship between the tasks is modeled with a single inter-task correlation matrix. We extend this by modeling the inter-task covariance by the sum of a set of covariance functions, weighted by a set of inter-task correlation matrices. The rationale behind our method is that by using a set of covariance functions and inter-task correlation matrices, more free parameters are introduced in the model and, subsequently, the model becomes more expressive and can better learn the complex relationship between the biochemicals and the spectra.

Let $\boldsymbol{X}={(\mathbf{x}_1,...,\mathbf{x}_N)}^T$ be the \textit{N} distinct inputs, $\mathbf{Y}_{N \times M}$ be the corresponding \textit{M} task  to learn and $\boldsymbol{y}= vec(\mathbf{Y}) = (y_{11},...,y_{N1}...,y_{1M},...,y_{NM})^T$ such that $y_{il}$ is the output of $\mathbf{x}_i$ on $l^\text{th}$ task. The inter-task covariance is defined as 
\begin{eqnarray}
\label{eq:multitaskeq}
\left<f_l(\mathbf{x}_i),f_k(\mathbf{x}_j)\right> \, &=& \, \mathbf{P}^{f}_{l,k} \, k_1^x( \mathbf{x}_i, \mathbf{x}_j) + \mathbf{Q}^{f}_{l,k} \, k_2^x( \mathbf{x}_i, \mathbf{x}_j), 
\end{eqnarray}

where $\mathbf{P}^{f}$ and $\mathbf{Q}^{f}$ are the $M \times M$ positive semi-definite task correlation matrices for the two covariance functions $k_1(\mathbf{x},\mathbf{x'})$ and $k_2(\mathbf{x},\mathbf{x'})$, respectively.  The $(l,k)^\text{th}$ element of $\mathbf{P}^f$ and $\mathbf{Q}^f$ represent the correlation between the $l^\text{th}$ and the $k^\text{th}$ tasks relating to the covariance functions respectively. Then, the output is modeled as
\begin{eqnarray}
y_{il} &=&\mathcal{N}(f_l(\mathbf{x}_i), {\sigma}_l^2 ),
\label{eq:multitaskeq1}
\end{eqnarray}

where ${\sigma}_l^2$ is the noise variance in task $l$.

\subsection{Inference and learning hyperparameters}
The standard Gaussian process formulation can be used to make inference on this model~\cite{rasmussen2005}. The mean of the predictive distribution at point $\mathbf{x}_*$ for the $l^\text{th}$ task, $\bar{f_l}(\mathbf{x}_*)$, is given by 
\begin{eqnarray}
\label{eq:multitask_pred1}
\bar{f_l}(\mathbf{x}_*) &=& {(\mathbf{p}^f_l  \otimes \mathbf{k}^x_{1*} + \mathbf{q}^f_l  \otimes \mathbf{k}^x_{2*} )}^T {\Sigma}^\text{-1} \mathbf{y}, 
\end{eqnarray}
where
\begin{eqnarray}
\Sigma &=&  \mathbf{P}^{f}  \otimes \mathbf{K}^x_1 + \mathbf{Q}^{f}  \otimes \mathbf{K}^x_2 + \mathbf{D}  \otimes \mathbf{I} 
\label{eq:multitask_pred2}
\end{eqnarray}

and, $\mathbf{p}^f_l$ and $\mathbf{q}^f_l $ are the $l^\text{th}$ row or column of $\mathbf{P}^{f}$ and $\mathbf{Q}^{f}$ respectively. $\mathbf{k}^x_{1*}$ and $\mathbf{k}^x_{2*}$ are the vectors of covariances between the point and the training points using two covariance functions respectively. $\mathbf{K}^x_1$ and $\mathbf{K}^x_2$ are the $MN \times MN$ matrices of covariance between all pairs of training points using two covariance functions respectively. $D$ is a $M \times M$ matrix with noise variance in each task $l$ as its $l^\text{th}$ diagonal element and $I$ is a $N \times N$ identity matrix.

The matrices $\mathbf{P}^f$ and $\mathbf{Q}^f$ should be constrained to be positive semi-definite. For this purpose, similar to in \cite{rakitsch2013}, each correlation matrix can be modeled as  $a_0^2 + \sum_{i=1}^k \mathbf{b}_i^T \mathbf{b}_i$, where $k$ is the rank of the correlation matrix, $a$ is a scalar and $\mathbf{b}_i$ for $i= 1...k$ are column vectors of length equal to the dimension of the correlation matrix. The rank, $k$, is manually set, while $a$ and $\mathbf{b}_i$ are learned, along with the hyperparameters of $\mathbf{k}^x_{1*}$ and $\mathbf{k}^x_{2*}$, by minimizing the negative log likelihood
\begin{equation}
\label{eq:multitask_nlml}
\log(\mathbf{y} | \mathbf{X} ) = - \frac{1}{2} \mathbf{y}^T {\Sigma}^\text{-1}\mathbf{y} - \frac{1}{2} \log |\Sigma| - \frac{NM}{2} \log(2\pi).
\end{equation}

Extending this formulation to include more than two covariance functions to make the relationship between the task more flexible, or to have correlated noise, as in \cite{rakitsch2013}, is straightforward. However, with addition of every new covariance funtion or addition of correlated noise, more parameters have to be learned and more covariance matrices need to be computated. This could lead to over-fitting of the parameters and increased computational overhead. Hence, in this paper as a proof-of-concept, we focus on the case with two covariance functions with uncorrelated noise.

\section{Experiments}
\label{sec:experiments}

\begin{table*}[!htb]
\caption{Performance measured by the mean and the standard deviation of the predictive $r^2$ over 50 independent trials.}
\label{table}
\centering
\begin{tabular}{llllll}
\toprule
Method & Chlorophyll-a & Chlorophyll-b & Nitrogen & Carbon\tabularnewline
\midrule
GP (SE) & 0.5972 ($\pm$0.112) & 0.4986 ($\pm$0.136) & 0.4515 ($\pm$0.201) & 0.4535 ($\pm$0.162)\tabularnewline
GP (NN) & 0.6405 ($\pm$0.114) & 0.5329 ($\pm$0.129) & 0.5331 ($\pm$0.132) & 0.5162 ($\pm$0.168)\tabularnewline
GP (SUM) & 0.6238 ($\pm$0.115) & 0.5203 ($\pm$0.133) & 0.5166 ($\pm$0.132) & 0.5192 ($\pm$0.164)\tabularnewline
MTGP-SC (SE) & 0.6284 ($\pm$0.079) & 0.5805 ($\pm$0.119) & 0.5833 ($\pm$0.130) & 0.5518 ($\pm$0.150)\tabularnewline
MTGP-SC (NN) & 0.6263 ($\pm$0.093) & 0.5763 ($\pm$0.118) & 0.6283 ($\pm$0.112) & 0.6015 ($\pm$0.151)\tabularnewline
MTGP-SC (SUM) & 0.6591 ($\pm$0.109) & 0.6343 ($\pm$0.146) & 0.6686 ($\pm$0.124) & 0.6231 ($\pm$0.153)\tabularnewline
MTGP-SN (SE) & 0.6427 ($\pm$0.083) & 0.5474 ($\pm$0.133) & 0.4621 ($\pm$0.148) & 0.5021 ($\pm$0.140)\tabularnewline
MTGP-SN (NN) & 0.6430 ($\pm$0.098) & 0.5803 ($\pm$0.151) & 0.5895 ($\pm$0.131) & 0.5927 ($\pm$0.155)\tabularnewline
MTGP-SN (SUM) & 0.6796 ($\pm$0.100) & \textbf{0.6575} ($\pm$0.147) & 0.6629 ($\pm$0.109) & 0.6000 ($\pm$0.160)\tabularnewline
MTGP-COMP (SE, NN)* & \textbf{0.6869} ($\pm$0.116) & 0.6177 ($\pm$0.142) & \textbf{0.7262} ($\pm$0.107) & \textbf{0.6569} ($\pm$0.142)\tabularnewline
\bottomrule
*proposed method.
\end{tabular}
\end{table*}

We present experiments using two datasets. We examine the common situation where it is harder or more expensive to obtain analysis about some biochemical quantities, leading to the datasets having few ground truth for some quantities while having plenty ground truth for other quantities. The methods compared are multitask learning with composite covariance function (MTGP-COMP), multitask learning with shared covariance (MTGP-SC)~\cite{bonilla2008}, multitask learning with structured noise (MTGP-SN)~\cite{rakitsch2013} and single-task Gaussian Process (GP).

\subsection{ Datasets} 

The first dataset contains 103 reflectance spectra of sediments containing algal bio-films, and the contents of the chlorophyll-a and the chlorophyll-b in \si{\micro\gram\per\square\centi\metre}. The dataset was acquired by Murphy et al.~\cite{murphy2005} from two mudflats, each of an area about \SI{500}{\square\metre}, in Sydney, Australia. The reflectance spectra is measured in visible and near infrared region (350-1050 \si{\nano\metre} at \SI{1}{\nano\metre} interval).  The second dataset contains 54 reflectance spectra of foliage and the corresponding nitrogen and carbon contents of the samples, measured in terms of percentage dry foliage weight, collected as part of a field campaign by The National Ecological Observatory Network (NEON)\footnote{National Ecological Observatory Network. 2015.  Available on-line  http://data.neoninc.org/ from National Ecological Observatory Network, Boulder, CO, USA.}. It contains visible to shortwave infrared spectra (350-2050 \si{\nano\metre} at \SI{1}{\nano\metre} interval). It also contains an airborne hyperspectral image of a \SI{250}{\metre} $\times$ \SI{250}{\metre} test area. 

\subsection{Methodology}
Out of the two biochemical quantities in each dataset, one was chosen to be the primary quantity and the other to be the secondary quantity (similar to \cite{schneider2014,alvarez2009}). We assume that there are few training examples for the primary quantities while a large number of training examples for the secondary quantities. We evaluate the efficacy of the multitask method by estimating the missing values of the primary quantities of the samples for which the values of the secondary quantities are available. We repeat the same process by making the primary quantity in the first step as the secondary quantity and the secondary quantity in the first step as the primary quantity.

The examples in the datasets were randomly separated into the training and the test sets. The test set contained one-third of the ground truth instances of the primary quantity. All the instances of the secondary quantity and the remaining two-third instances of the primary quantity were included in the training set. Eighty percent of the training set was used to train models with different rank approximations of the correlation matrices in the multitask models. Hyperparameters of the covariance functions and the correlation matrices were learned by minimizing the log likelihood using quasi-Newton method. Five trials of this optimization were performed using random initial guesses to prevent local minima. The ranks of the correlation matrices and the hyperparameters that produced the best $r^2$ in predicting the remaining twenty percent of the training examples were chosen as final correlation matrix ranks and hyperparameters. Using them, predictions were made on the testing set and the performance was measured by $r^2$ value. This procedure was repeated for 50 independent trials, and the mean and the standard deviation of the measured $r^2$ value are reported.

\subsection{Results}

\begin{figure}

\centering
	 \subfloat[][RGB Image]{\includegraphics[scale=0.95]{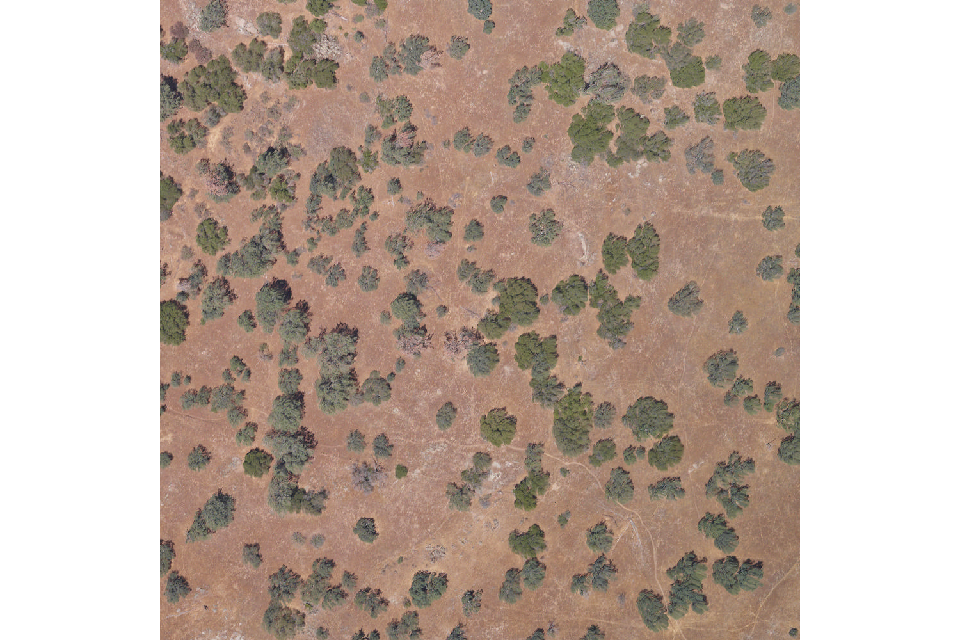}\label{i1}} \\
     \subfloat[][Predicted Nitrogen Contents (\% leaf weight)]{\includegraphics[scale=0.95]{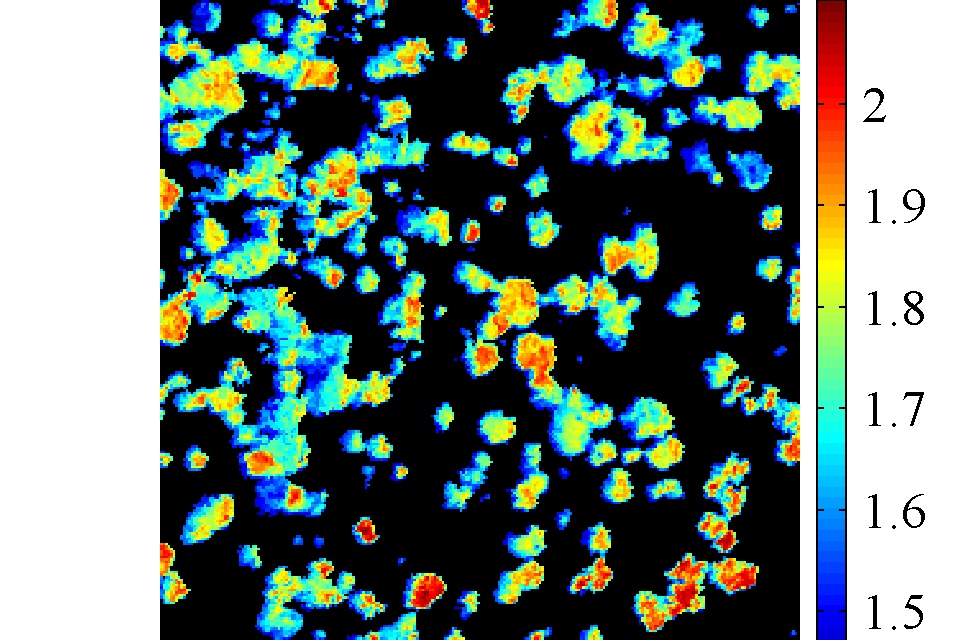}\label{i2}} \\
     \subfloat[][Predicted Carbon Contents (\% leaf weight)]{\includegraphics[scale=0.95]{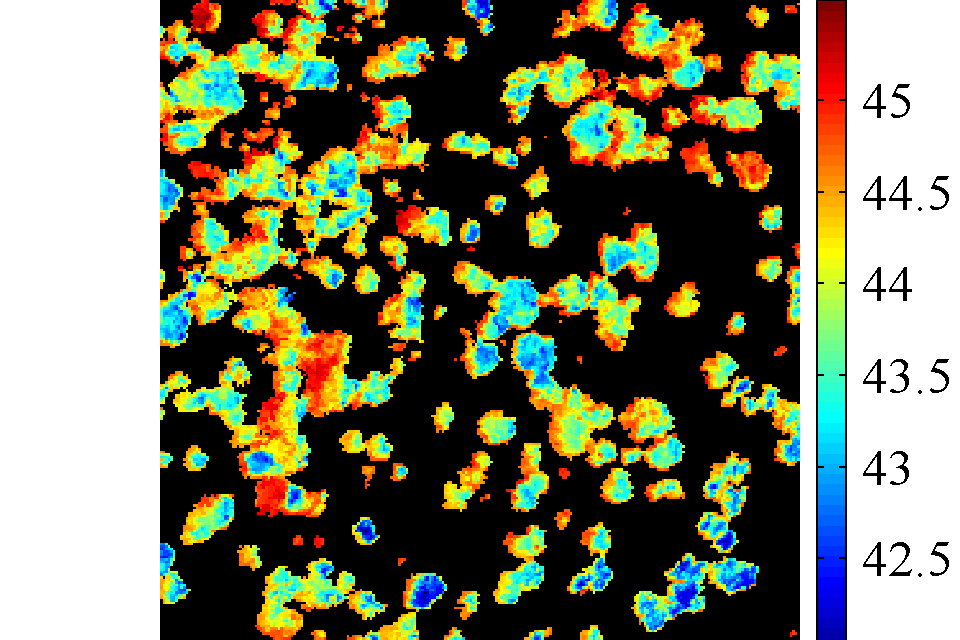}\label{i3}} 
\caption{The proposed method applied to a \SI{250}{\metre} $\times$ \SI{250}{\metre} airborne hyperspectral image.}
\label{fig:images}
\end{figure}

Table \ref{table} summarizes the results of the experiment on both the datasets. The covariance functions used with each method is given in the parenthesis alongside method's name. All the covariance functions used were isotropic. SE stands for the squared exponential covariance function, NN stands for the neural network covariance function and SUM stands for the sum covariance function formed by summing the squared exponential and the neural network covariance functions.  Both squared exponential (SE) and neural network (NN) covariance functions were used with the proposed method (MTGP-COMP). For illustration, fig. \ref{fig:images} shows the nitrogen and the carbon contents prediction map generated from the test hyperspectral image by the MTGP-COMP model (one out of the 50 trials). As pre-processing, the water absorption bands were removed and the sampling wavelengths of the ground spectra and the image spectra were matched using linear interpolation. Non-vegetation pixels (pixels with Normalized Difference Vegetation Index less than 0.3) have been masked out.

\section{Discussion}
\label{sec:discussion}
The proposed method presented the best result for three out of four biochemicals. This demonstrates that using a combination of covariance functions and task correlation matrices can produce more flexible models yielding more accurate results, compared to the previous multitask methods. Also, all multitask methods performed better than the single-task GP for all biochemicals, confirming the hypothesis that multitask learning improves the biochemical content prediction when few training data are available. The difference in the mean $r^2$ of the single task GP and the proposed method is quite significant. However, the standard deviation of $r^2$ is fairly large for all the methods. It is probably due to the limited number of training examples. The standard deviation of $r^2$ is consistent between all the methods, indicating that the mean $r^2$ value is good representative of the trend in the performance. The learning in the current implementation of the proposed method is slow and not scalable to cases where there are multiple secondary quantities. As future work, sparse Gaussian processes~\cite{snelson2005} could be used with the proposed method to reduce the computational complexity.

\bibliographystyle{IEEEbib}
\bibliography{whispers1}

\end{document}